\documentclass[10pt,twocolumn,letterpaper]{article}

\usepackage{cvpr}
\usepackage{times}
\usepackage{epsfig}
\usepackage{graphicx}
\usepackage{tabularx}
\usepackage{amsmath,mathtools}
\usepackage{amssymb}
\usepackage[inline]{enumitem}
\usepackage{multirow}
\usepackage{subcaption}
\usepackage[dvipsnames]{xcolor}
\usepackage{booktabs}
\usepackage{algpseudocode,algorithm,algorithmicx}
\usepackage{xspace}
\usepackage{interval}
\usepackage{bm}
\usepackage[pagebackref=true,breaklinks=true,letterpaper=true,colorlinks,citecolor=ForestGreen, bookmarks=false]{hyperref}

\newcommand{\system}{AdaFrame\xspace}

\newcommand{\E}{\mathbb{E}}
\newcommand{\Ea}[1]{\E\left[#1\right]}
\newcommand{\Eb}[2]{\E_{#1}\left[#2\right]}
\newcommand{\Vpi}{{V_t}}

\newcommand{\delay}{i}
\DeclarePairedDelimiter{\norm}{\|}{\|}
\DeclareMathOperator*{\minimize}{minimize}
\newcommand{\grad}{\nabla}
\newcommand{\gradth}{\grad_{\Theta}}
\newcommand{\pith}{\pi_{\theta}}
\newcommand{\given}[1][]{\:#1\vert\:}

\newcommand{\ra}[1]{\renewcommand{\arraystretch}{#1}}
\newcommand{\anet}{{\scshape ActivityNet}\xspace}
\newcommand{\fcvid}{{\scshape FCVID}\xspace}

\definecolor{Gray}{gray}{0.85}
\newcolumntype{g}{>{\columncolor{Gray}} c}
\newcommand{\arrto}{$\,\to\,$}
\cvprfinalcopy %

\ifcvprfinal\fi
\begin{document}

\title{AdaFrame: Adaptive Frame Selection for Fast Video Recognition}
\newcommand\blfootnote[1]{%
  \begingroup
  \renewcommand\thefootnote{}\footnote{#1}%
  \addtocounter{footnote}{-1}%
  \endgroup
}
\author{Zuxuan Wu$^{1*}$, Caiming Xiong$^{2\dagger}$, Chih-Yao Ma$^{3}$, Richard Socher$^{2}$, Larry S. Davis$^{1}$ \\
$^{1}$ University of Maryland, $^{2}$ Salesforce Research, $^{3}$ Georgia Institute of Technology}

\maketitle

\begin{abstract}
\blfootnote{$^{*}$ Most of the work is done when the author was an intern at Salesforce.}
\blfootnote{$^{\dagger}$ Corresponding author.}
We present AdaFrame, a framework that adaptively selects relevant frames on a per-input basis for fast video recognition. AdaFrame contains a Long Short-Term Memory network augmented with a global memory that provides context information for searching which frames to use over time. Trained with policy gradient methods, AdaFrame generates a prediction, determines which frame to observe next, and computes the utility, i.e., expected future rewards, of seeing more frames at each time step. At testing time, AdaFrame exploits predicted utilities to achieve adaptive lookahead inference such that the overall computational costs are reduced without incurring a decrease in accuracy. Extensive experiments are conducted on two large-scale video benchmarks, FCVID and ActivityNet. AdaFrame matches the performance of using all frames with only 8.21 and 8.65 frames on FCVID and ActivityNet, respectively. We further qualitatively demonstrate learned frame usage can indicate the difficulty of making classification decisions; easier samples need fewer frames while harder ones require more, both at instance-level within the same class and at class-level among different categories. 
\end{abstract}

\section{Introduction}
The explosive increase of Internet videos, driven by the ubiquity of mobile devices and sharing activities on social networks, is phenomenal: around 300 hours of video are uploaded to YouTube every minute of every day! Such growth demands effective and scalable approaches that can recognize actions and events in videos automatically for tasks like indexing, summarization, recommendation, \etc. Most existing work focuses on learning robust video representations to boost accuracy~\cite{C3D,wang2017non,DBLP:conf/nips/SimonyanZ14,WangXWQLTV16}, while limited effort has been devoted to improving efficiency~\cite{wu2018compressed,ZhangWWQW16}.

State-of-the-art video recognition frameworks rely on the aggregation of prediction scores from uniformly sampled frames~\footnote{Here, we use frame as a general term, and it can be in the forms of a single RGB image, stacked RGB images (snippets), and stacked optical flow images.}, if not every single frame~\cite{ng2015beyond}, during inference. While uniform sampling has been shown to be effective~\cite{DBLP:conf/nips/SimonyanZ14,WangXWQLTV16,wang2017non}, the analysis of even a single frame is still computationally expensive due to the use of high-capacity backbone networks such as ResNet~\cite{he2015deep}, ResNext~\cite{xie2017aggregated}, InceptionNet~\cite{szegedy2017inception}, \etc. On the other hand, uniform sampling assumes information is evenly distributed over time, which could therefore incorporate noisy background frames that are not relevant to the class of interest. 

It is also worth noting that the difficulty of making recognition decisions relates to the category to be classified---one frame might be sufficient to recognize most static objects (\eg, ``dogs'' and ``cats'') or scenes (\eg, ``forests'' or ``sea'') while more frames are required to differentiate subtle actions like ``drinking coffee'' and ``drinking beer''. This also holds for samples even within the same category due to large intra-class variations. For example, a ``playing basketball'' event can be captured from multiple view points (\eg, different locations of a gymnasium), occur at different locations (\eg, indoor or outdoor), with different players (\eg, professionals or amateurs). As a result, the number of frames required to recognize the same event are different.

\begin{figure}[t!]
\centering
\includegraphics[width=0.83\columnwidth]{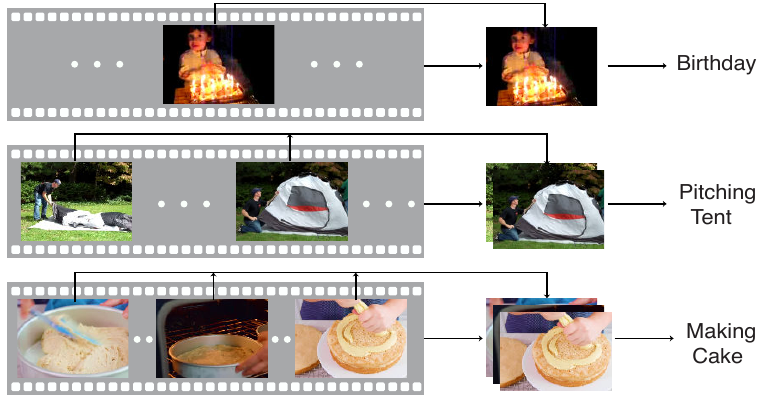}
\caption{ \textbf{A conceptual overview of our approach}. \system aims to select a small number of frames to make correct predictions conditioned on different input videos so as to reduce the overall computational cost.} 
\vspace{-0.2in}
\label{fig:teaser}
\end{figure}

With this in mind, to achieve efficient video recognition, we explore how to automatically adjust computation within a network on a per-video basis such that---conditioned on different input videos, a small number of informative frames are selected to produce correct predictions (See Figure~\ref{fig:teaser}). 
However, this is a particularly challenging problem, since videos are generally weakly-labeled for classification tasks, one annotation for a whole sequence, and there is no supervision informing which frames are important. Therefore, it is unclear how to effectively explore temporal information over time to choose which frames to use, and how to encode temporal dynamics in these selected frames.

In this paper, we propose \system, a Long Short-Term Memory (LSTM) network augmented with a global memory, to learn how to adaptively select frames conditioned on inputs for fast video recognition. In particular, a global memory derived from representations computed with spatially and temporally downsampled video frames is introduced to guide the exploration over time for learning frame usage policies. The memory-augmented LSTM serves as an agent interacting with video sequences; at a time step, it examines the current frame, and with the assistance of global context information derived by querying the global memory, generates a prediction, decides which frame to look at next and calculates the utility of seeing more frames in the future. During training, \system is optimized using policy gradient methods with a fixed number of steps to maximize a reward function that encourages predictions to be more confident when observing one more frame. At testing time, \system is able to achieve adaptive inference conditioned on input videos by exploiting the predicted future utilities that indicate the advantages of going forward.

We conduct extensive experiments on two large-scale and challenging video benchmarks for generic video categorization (\fcvid~\cite{TPAMI-fcvid}) and activity recognition (\anet~\cite{caba2015activitynet}). \system offers similar or better accuracies measured in mean average precision over the widely adopted uniform sampling strategy, a simple yet strong baseline, on \fcvid and \anet respectively, while requiring $58.9\%$ and $63.3\%$ fewer computations on average, going as high as savings of $90.6\%$. \system also outperforms by clear margins alternative methods~\cite{yeung2016end,fan18ijcai} that learn to select frames. We further show that, among other things, frame usage is correlated with the difficulty of making predictions---different categories produce different frame usage patterns and instance-level frame usage within the same class also differs. These results corroborate that \system can effectively learn to generate frame usage policies that adaptively select a small number of relevant frames for classification for each input video.  

\section{Related Work}
\noindent\textbf{Video Analysis}. 
Extensive studies have been conducted on video recognition~\cite{wu2017deeplearning}. Most existing work focuses on extending 2D convolution to the video domain and modeling motion information in videos~\cite{DBLP:conf/nips/SimonyanZ14,wang2017non,WangXWQLTV16,xie2018rethinking,C3D}. Only a few methods consider efficient video classification~\cite{ZhangWWQW16,wu2018compressed,zolfaghari2018eco,su2016leaving,TMM:RealTimeEvent}. However, these approaches perform mean-pooling of scores/features from multiple frames, either uniformly sampled or decided by an agent, to classify a video clip. In contrast, we focus on selecting a small number of relevant frames, whose temporal relations are modeled by an LSTM, on a per-video basis for efficient recognition. Note that our framework is also applicable to 3D CNNs; the inputs to our framework can be easily replaced with features from stacked frames. A few recent approaches attempt to reduce computation cost in videos by exploring similarities among adjacent frames~\cite{zhu2017deep,pan2018recurrent}, while our goal is to selectively choose relevant frames based on inputs.

Our work is more related to \cite{yeung2016end} and \cite{fan18ijcai} that choose frames with policy search methods~\cite{deisenroth2013survey}. Yeung \etal introduce an agent to predict whether to stop and where to look next through sampling from the whole video for action detection~\cite{yeung2016end}. For detection, ground-truth temporal boundaries are available, providing strong feedback about whether viewed frames are relevant. In the context of classification, there is no such supervision, and thus directly sampling from the entire sequence is difficult.  To overcome this issue, Fan \etal propose to sample from a predefined action set deciding how many steps to jump~\cite{fan18ijcai}, which reduces the search space but sacrifices flexibility. In contrast, we introduce a global memory module that provides context information to guide the frame selection process. We also decouple the learning of frame selection and when to stop, exploiting predicted future returns as stop signals.

\begin{figure*}[t!]
\begin{center}
   \includegraphics[width=0.83\linewidth]{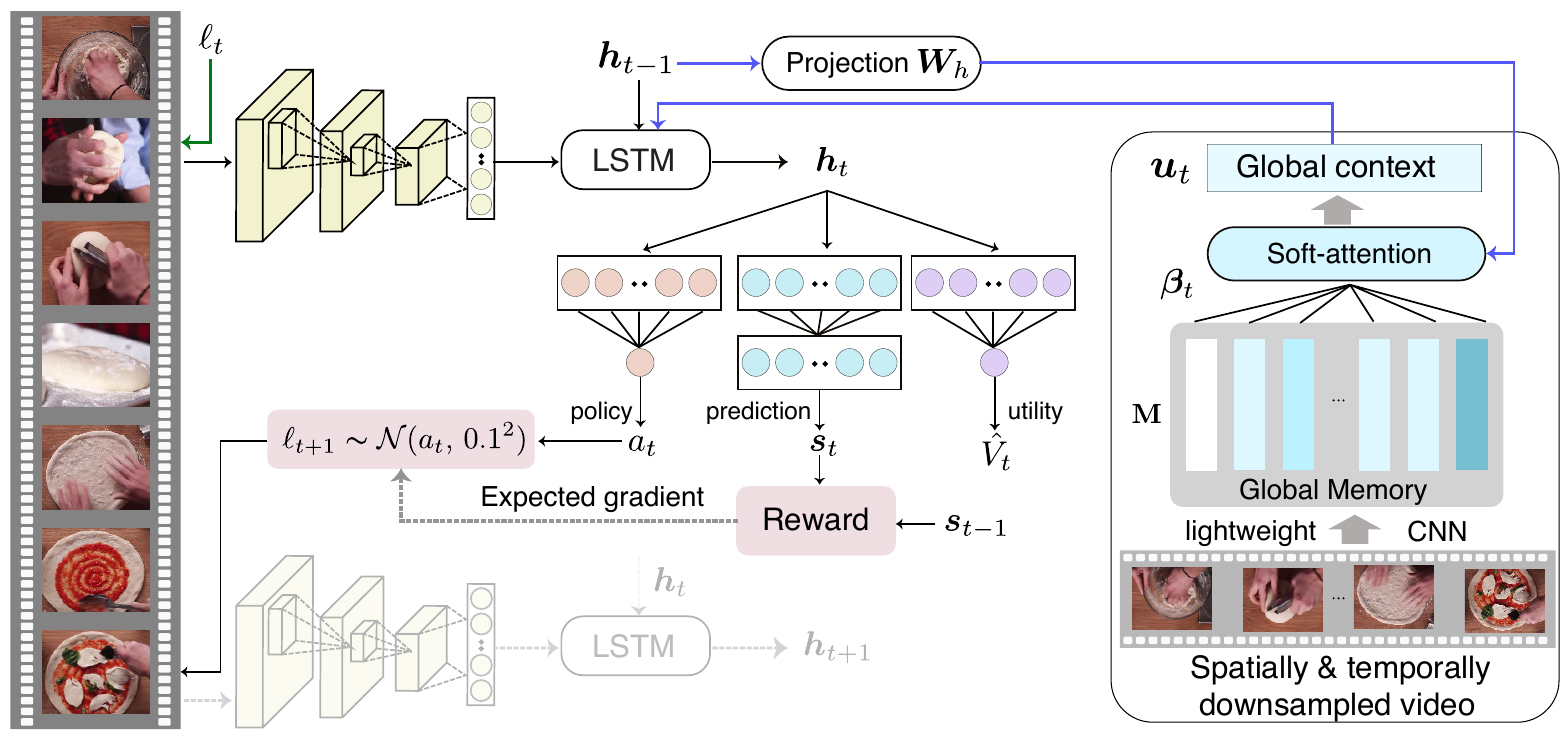}
\end{center}
\vspace{-0.3in}
   \caption{\textbf{An overview of the proposed framework}. A memory-agumented LSTM serves as an agent, interacting with a video sequence. At each time step, it takes features from the current frame, previous states, and a global context vector derived from a global memory to generate the current hidden states. The hidden states are used to produce a prediction, decides where to look next and calculates the utility of seeing more frames in the future. See texts for more details.}
\label{fig:framework}
\end{figure*}

\vspace{0.05in}
\noindent\textbf{Adaptive Computation}. Our work also relates to adaptive computation to achieve efficiency by deciding whether to stop inference based on the confidence of classifiers. The idea dates back to cascaded classifiers~\cite{viola2004robust} that quickly reject easy negative sub-windows for fast face detection. Several recent approaches propose to add decision branches to different layers of CNNs to learn whether to exit the model~\cite{teerapittayanon2016branchynet,huang2017multi,mcgill2017deciding,figurnov2017spatially}. Graves introduce a halting unit to RNNs to decide whether computation should continue~\cite{graves2016adaptive}. Related are also~\cite{wang2017skipnet,veit2018convolutional,wu2018blockdrop,najibi2018autofocus,gao2018dynamic} that learn to drop layers in residual networks or learn where to look in images conditioned on inputs. In this paper, we focus on adaptive computation for videos to adaptively select frames rather than layers/units in neural networks for fast inference.

\section{Approach}
Our goal is, given a testing video, to derive an effective frame selection strategy that produces a correct prediction while using as few frames as possible. To this end, we introduce \system, a memory-augmented LSTM (Section~\ref{sec:mlstm}), to explore the temporal space of videos effectively with the guidance of context information from a global memory. \system is optimized to choose which frames to use on a per-video basis, and to capture the temporal dynamics of these selected frames. Given the learned model, we perform adaptive lookahead inference (Section~\ref{sec:inference}) to accommodate different computational needs through exploring the utility of seeing more frames in the future.

\subsection{Memory-augmented LSTM}
\label{sec:mlstm}
The memory-augmented LSTM can be seen as an agent that recurrently interacts with a video sequence of $T$ frames, whose representations are denoted as $\{{\bm v}_1,{\bm v}_2,\ldots,{\bm v}_T\}$. More formally, the LSTM, at the $t$-th time step, takes features of the current frame ${\bm v}_t$, previous hidden states ${\bm h}_{t-1}$ and cell outputs ${\bm c}_{t-1}$, as well as a global context vector ${\bm u}_t$ derived from a global memory ${\bf M}$ as its inputs, and produces the current hidden states ${\bm h}_{t}$ and cell contents ${\bm c}_{t}$:
\begin{equation}
\label{eq:lstm}
{\bm h}_t, \, {\bm c}_t = \texttt{LSTM}([{\bm v}_t, {\bm u}_t], \, {\bm h}_{t-1}, \, {\bm c}_{t-1}),
\end{equation}
where ${\bm v}_t$ and ${\bm u}_t$ are concatenated. The hidden states ${\bm h}_t$ are further input into a prediction network $f_p$ for classification, and the probabilities are used to generate a reward ${r}_t$ measuring whether the transition from the last time step brings information gain. Furthermore, conditioned on the hidden states, a selection network $f_s$ decides where to look next, and a utility network $f_u$ calculates the advantage of seeing more frames in the future. Figure~\ref{fig:framework} gives an overview of the framework. In the following, we elaborate detailed components in the memory-augmented LSTM.

\vspace{0.05in}
\noindent\textbf{Global memory}. The LSTM is expected to make reliable predictions and explore the temporal space to select frames guided by rewards received. However, learning where to look next is difficult due to the huge search space and limited capacity of hidden states~\cite{collins2016capacity,yogatama2018memory} to remember input history. Therefore, for each video, we introduce a global memory to provide context information, which consists of representations of spatially and temporally downsampled frames, ${{\bf M}} = [{\bm v}_1^s,{\bm v}_2^s,\ldots,{\bm v}_{T_d}^s]$. Here, $T_d$ denotes the number of frames ($T_d < T$), and the representations are computed with a lightweight network using spatially downsampled inputs (more details in Sec.~\ref{sec:exp}). This is to ensure the computational overhead of the global memory is small. As these representations are computed frame by frame without explicit order information, we further utilize positional encoding~\cite{vaswani2017attention} to encode positions in the downsampled representations. To obtain global context information, we query the global memory with the hidden states of the LSTM to get an attention weight for each element in the memory:
\begin{align*}
z_{t,j} =  ({\bm W}_{h} {\bm h}_{t-1})^\top PE({\bm v}_j^s), \quad {\bm \beta}_{t} = &  \texttt{Softmax}( {\bm z}_t ),
\end{align*}
where $\bm{W}_h$ maps hidden states to the same dimension as the $j$-th downsampled feature ${\bm v}_j^s$ in the memory, $PE$ denotes the operation of adding positional encoding to features, and $\bm{\beta}_t$ is the normalized attention vector over the memory.
We can further derive the global context vector as the weighted average of the global memory: $\bm{u}_t = {\bm \beta}_{t}^\top{\bf M}$. The intuition of computing a global context vector with soft-attention as inputs to the LSTM is to derive a rough estimate of the current progress based on features in the memory block, serving as global context to assist the learning of which frame in the future to examine.

\vspace{0.05in}
\noindent\textbf{Prediction network}. The prediction network $f_p({\bm h}_t; {\bm W}_p$) parameterized by weights ${\bm W}_p$ maps the hidden states ${\bm h}_t$ to outputs ${\bm s}_t \in {\mathbb R}^C$ with one fully-connected layer, where $C$ is the number of classes. In addition, ${\bm s}_t$ is further normalized with $\texttt{Softmax}$ to produce probability scores for each class. 
The network is trained with cross-entropy loss using predictions from the last time step $T_e$: 
\begin{align}
\label{eqn:xe}
\mathcal{L}_{cls}({\bm W}_p) = -\sum_{c=1}^{C}{{\bm y}^{c}\,\log({\bm s}^{c}_{T_{e}})}, 
\end{align}
where ${\bf y}$ is a one-hot vector encoding the label of the corresponding sample. In addition, we constrain $T_e \ll T$, since we wish to use as few frames as possible.

\vspace{0.05in}
\noindent\textbf{Reward function}. Given the classification scores $\bm{s}_t$ of the $t$-th time step, a reward is given to evaluate whether the transition from the previous time step is useful---observing one more frame is expected to produce more accurate predictions. Inspired by~\cite{ma2016learning}, we introduce a reward function that forces the classifier to be more confident when seeing additional frames, taking the following form (when $t>1$):
\begin{align}
r_t = 
    \max \{0, \, m_{t} \,- \max_{t'\in\interval{0}{t-1}}m_{t'} \}.   
\label{eqn:reward}
\end{align}
Here, $m_t = \bm{s}_t^{{gt}} - \max\{\bm{s}_t^{c'} | c' \neq {gt} \}$ is the margin between the probability of the ground-truth class (indexed by $gt$) and the largest probabilities from other classes, pushing the score of the ground-truth class to be higher than other classes by a margin. And the reward function in Eqn.~\ref{eqn:reward} encourages the current margin to be larger than historical ones to receive a positive reward, which demands that the confidence of the classifier increases when seeing more frames. Such a constraint acts as a proxy to measure if the transition from the last time step brings additional information for recognizing target classes, as there is no supervision providing feedback about whether a single frame is informative. 

\vspace{0.05in}
\noindent\textbf{Selection network}. The selection network $f_s$ defines a policy with a Gaussian distribution using fixed variance, to decide which frame to observe next, using hidden states ${\bm h}_t$ that contain information of current inputs and historical context. In particular, the network, parameterized by ${\bm W}_s$, transforms the hidden states to a 1-dimensional output $  f_s({\bm h}_t; {\bm W}_s) = a_t = \texttt{sigmoid}({\bm W}_s^\top {\bm h}_t)$, as the mean of the location policy. Following~\cite{mnih2014recurrent}, during training, we sample from the policy $\ell_{t+1} \thicksim \pi( \cdot | \bm{h}_t) = \mathcal{N}(a_t,\,0.1^{2})$, and at testing time, we directly use the output as the location. We also clamp $\ell_{t+1}$ to be in the interval of $\interval{0}{1}$, so that it can be further transfered to a frame index multiplying by the total number of frames. It is worth noting that at the current time step, the policy searches through the entire time horizon and there is no constraint; it can not only jump forward to seek future informative frames but also go back to re-examine past information. We train the selection network to maximize the expected future reward:
\begin{align}
\label{eqn:accureward}
J_{sel}({\bm W}_s) = \Eb{\ell_t \thicksim \pi(\cdot | \bm{h}_t; {\bm W}_s)}{\sum_{t=0}^{T_e}r_t}.
\end{align}

\vspace{0.05in}
\noindent\textbf{Utility network}. The utility network, parameterized by ${\bm W}_u$, produces an output $f_u({\bm h}_t;{\bm W}_u) = \hat{V_t} = {\bm W}_u^\top{\bm h}_t$ using one fully-connected layer. It serves as a critic to provide an approximation of expected future rewards from the current state, which is also known as the value function~\cite{sutton1998reinforcement}:
\begin{align}
\Vpi & = \Eb{\substack{{\bm h}_{t+1:T_e},\\a_{t:T_e} } }{\sum_{\delay=0}^{T_e - t} \gamma^\delay r_{t+\delay}}, 
\end{align}
where $\gamma$ is the discount factor fixed to 0.9. The intuition is to estimate the value function $\Vpi$ derived from empirical rollouts with the network output $\hat{V_t}$ to update policy parameters in the direction of performance improvement. More importantly, by estimating future returns, it provides the agent with the ability to look ahead, measuring the utility of subsequently observing more frames. The utility network is trained with the following regression loss:
\begin{align}
\label{eqn:value}
L_{utl}({\bm W}_u) = \frac{1}{2}\norm{\hat{V_t} - \Vpi}_2.
\end{align}

\vspace{0.05in}
\noindent\textbf{Optimization}.
Combining Eqn.~\ref{eqn:xe}, Eqn.~\ref{eqn:accureward} and Eqn.~\ref{eqn:value}, the final objective function can be written as:
\begin{align*}
\minimize_{\Theta} \, \mathcal{L}_{cls} + \lambda\mathcal{L}_{utl} - \lambda J_{sel},
\end{align*}
where $\lambda$ controls the trade off between classification and temporal exploration and $\Theta$ denotes all trainable parameters.
Note that the first two terms are differentiable, and we can directly use back propagation with stochastic gradient descent to learn the optimal weights. Thus, we only discuss how to maximize the expected reward $J_{sel}$ in Eqn.~\ref{eqn:accureward}. Following~\cite{sutton1998reinforcement}, we derive the expected gradient of $J_{sel}$ as:

\begin{align}
\label{eqn:pg}
\gradth J_{sel} = 
\Ea{\sum_{t=0}^{T_e} (R_t - \hat{V_t})  \gradth \log \pith(\cdot \given \bm{h}_t)},
\end{align}
where $R_t$ denotes the expected future reward, and $\hat{V_t}$ serves as a baseline function to reduce variance during training~\cite{sutton1998reinforcement}. Eqn.~\ref{eqn:pg} can be approximated with Monte-Carlo sampling using samples in a mini-batch, and further back-propagated downstream for training.

\subsection{Adaptive Lookahead Inference}
\label{sec:inference}
While we optimize the memory-augmented LSTM for a fixed number of steps during training, we aim to achieve adaptive inference at testing time such that a small number of informative frames are selected conditioned on input videos without incurring any degradation in classification performance. Recall that the utility network is trained to predict expected future rewards, indicating the utility/advantage of seeing more frames in the future. Therefore, we explore the outputs of the utility network to determine whether to stop inference through looking ahead. A straightforward way is to calculate the utility $\hat{V_t}$ at each time step, and exit the model once it is less than a threshold. However, it is difficult to find an optimal value that works well for all samples. Instead, we maintain a running max of utility $\hat{V}^{max}$ over time for each sample, and at each time step, we compare the current utility $\hat{V}_{t}$ with the max value $\hat{V}_{t}^{max}$; if $\hat{V}_{t}^{max}$ is larger than $\hat{V}_{t}$ by a margin $\mu$ more than $p$ times, predictions from the current time step will be used as the final score and inference will be stopped. Here, $\mu$ controls the trade-off between computational cost and accuracy; a small $\mu$ constrains the model to make early predictions once the predicted utility begins to decrease while a large $\mu$ tolerates a drop in utility, allowing more considerations before classification. Further, we also introduce $p$ as a patience metric, which permits the current utility to deviate from the max value for a few iterations. This is similar in spirit to reducing learning rates on plateaus, which instead of intermediately decays learning rate waits for a few more epochs when the loss does not further decrease. 

Note that although the same threshold $\mu$ is used for all samples, comparisons made to decide whether to stop or not is based on the utility distribution of each sample independently, which is softer than comparing $\hat{V}_{t}$ with $\mu$ directly. 
One can add another network to predict whether to stop inference using the hidden states as in~\cite{yeung2016end,fan18ijcai}, however coupling the training of frame selection with learning a binary policy to stop makes optimization challenging, particularly with reinforcement learning, as will be shown in experiments. In contrast, we leverage the utility network to achieve adaptive lookahead inference.

\section{Experiments}
\subsection{Experimental Setup}
\label{sec:exp}
\noindent\textbf{Datasets and evaluation metrics}.  We experiment with two challenging large-scale video datasets, Fudan-Columbia Video Datasets (\fcvid)~\cite{TPAMI-fcvid} and \anet~\cite{caba2015activitynet}, to evaluate the proposed approach.  \fcvid consists of $91,223$ videos from YouTube with an average duration of $167$ seconds, manually annotated into 239 classes. These categories cover a wide range of topics, including scenes (\eg, {``river''}), objects (\eg, {``dog''}), activities (\eg, {``fencing''}), and complicated events (\eg, {``making pizza''}). The dataset is split evenly for training ($45,611$ videos) and testing ($45,612$ videos). \anet is an activity-focused large-scale video dataset, containing YouTube videos with an average duration of $117$ seconds. Here we adopt the latest release (version 1.3), which consists of around $20K$ videos belonging to $200$ classes. We use the official split with a training set of $10,024$ videos, a validation set of $4,926$ videos and a testing set of $5,044$ videos. Since the testing labels are not publicly available, we report performance on the validation set. We compute average precision (AP) for each class and use mean average precision (mAP) to measure the overall performance on both datasets. It is also worth noting that videos in both datasets are \emph{untrimmed}, for which efficient recognition is extremely critical given the redundant nature of video frames.

\vspace{0.05in}
\noindent\textbf{Implementation details}. We use a one-layer LSTM with $2,048$ and $1,024$ hidden units for \fcvid and \anet respectively.
To extract inputs for the LSTM, we decode videos at 1fps and compute features from the penultimate layer of a ResNet-101 model~\cite{he2015deep}.  The ResNet model is pretrained on ImageNet with a top-1 accuracy of $77.4\%$ and further finetuned on target datasets. To generate the global memory that provides context information, we compute features using spatially and temporally downsampled video frames with a lightweight CNN to reduce overhead. In particular, we lower the resolution of video frames to $112 \times 112$, and sample 16 frames uniformly. We use a pretrained MobileNetv2~\cite{sandler2018mobilenetv2} as the lightweight CNN, which achieves a top-1 accuracy of $52.3\%$ on ImageNet with downsampled inputs. We adopt PyTorch for implementation and leverage SGD for optimization with a momentum of $0.9$, a weight decay of $1e-4$ and a $\lambda$ of 1. We train the network for 100 epochs with a batch size of 128 and 64 for \fcvid and \anet, respectively. The initial learning rate is set to $1e-3$ and decayed by a factor of 10 every 40 epochs. For the patience $p$ during inference, it is set to 2 when $\mu<0.7$, and $K/2+1$ when $\mu=0.7$, where $K$ is number of time steps the model is trained for.

\subsection{Main Results}
\begin{table*}[t!]
\centering
\ra{1.}
\resizebox{1\linewidth}{!}{
\begin{tabular}{@{}c*{23}c@{}}
\toprule
& & \multicolumn{10}{c}{\fcvid} && \multicolumn{10}{c}{\anet}\\ 
\cmidrule{1-1} \cmidrule{3-12} \cmidrule{14-23}
Method && R8 & U8 && R10 & U10 && R25 & U25 && All && R8 & U8 && R10 & U10 && R25 & U25 && All \\
 \cmidrule{3-4} \cmidrule{6-7}  \cmidrule{9-10} \cmidrule{12-12} \cmidrule{14-15} \cmidrule{17-18} \cmidrule{20-21} \cmidrule{23-23}
AvgPooling && 78.3 & 78.4 && 79.0 & 78.9 && 79.7 & 80.0 && 80.2 &&   67.5 & 67.8 && 68.9 & 68.6 && 69.8 & 70.0 && 70.2 \\ 
LSTM 	   && 77.8 & 77.9 && 78.7 & 78.1 && 78.0 & 79.8 && 80.0 && 	68.7 & 68.8 && 69.8 & 70.4 && 69.9 & 70.8 && 71.0 \\ 
\cmidrule{1-1} \cmidrule{3-12} \cmidrule{14-23}
\multirow{2}{*}{\system} && 
\multicolumn{2}{c}{78.6} && \multicolumn{2}{c}{79.2} && \multicolumn{2}{c}{\textbf{80.2}} &&  && 
\multicolumn{2}{c}{69.5} && \multicolumn{2}{c}{70.4}  && \multicolumn{2}{c}{\textbf{71.5}} \\

&& 
\multicolumn{2}{c}{5 \arrto 4.92} && \multicolumn{2}{c}{8 \arrto 6.15} && \multicolumn{2}{c}{10 \arrto 8.21} &&  &&
\multicolumn{2}{c}{5 \arrto 3.8} && \multicolumn{2}{c}{8 \arrto 5.82} && \multicolumn{2}{c}{10 \arrto 8.65}\\
\bottomrule
\end{tabular}}
\vspace{-0.1in}
\caption{\textbf{Performance of different frame selection strategies on {\scshape{FCVID}} and {\scshape{ActivityNet}}.} R and U denote random and uniform sampling, respectively. We use $K$ \arrto $K'$ to denote the frame usage for \system, which uses $K$ frames during training and $K'$ frames on average when performing adaptive inference. See texts for more details.}
\label{tbl:results}
\end{table*}

\noindent\textbf{Effectiveness of learned frame usage}. We first optimize AdaFrame with $K$ steps during training and then at testing time we perform adaptive lookahead inference with $\mu=0.7$, allowing each video to see $K'$ frames on average while maintaining the same accuracy as viewing all $K$ frames. We compare \system with the following alternative methods to produce final predictions during testing: 
\begin{enumerate*}[label=(\arabic*)]
\item \textsc{AvgPooling}, which simply computes a prediction for each sampled frame and then performs a mean pooling over frames as the video-level classification score;
\item \textsc{LSTM}, which generates predictions using hidden states from the last time step of an LSTM.
\end{enumerate*}
We also experiment with different number of frames ($K + \Delta$) used as inputs for \textsc{AvgPooling} and \textsc{LSTM}, which are sampled either uniformly (U) or randomly (R). Here, we use $K$ for \system while $K + \Delta$ for other methods to offset the additional computation cost incurred, which will be discussed later. Table~\ref{tbl:results} presents the results. We observe \system achieves better results than \textsc{AvgPooling} and \textsc{LSTM} whiling using fewer frames under all settings on both datasets. In particular, \system achieves an mAP of $78.6\%$, and $69.5\%$ using an average of 4.92 and 3.8 frames on \fcvid and \anet respectively. These results, requiring 3.08 and 4.2 fewer frames, are better than \textsc{AvgPooling} and \textsc{LSTM} with 8 frames and comparable with their results with 10 frames. It is also promising to see that \system can match the performance of using all frames with only 8.21 and 8.65 frames on \fcvid and \anet. This verifies that \system can indeed learn to derive frame selection policies while maintaining the same accuracies.

In addition, the performance of random sampling and uniform sampling for \textsc{AvgPooling} and \textsc{LSTM} are similar and \textsc{LSTM} is worse than \textsc{AvgPooling} on \fcvid, possibly due to the diverse set of categories incur significant intra-class variations. Note that although \textsc{AvgPooling} is simple and straightforward, it is a very strong baseline and has been widely adopted during testing for almost all CNN-based approaches due to its strong performance.
 
\vspace{0.05in}
\noindent\textbf{Computational savings with adaptive inference}. We now discuss computational savings of \system with adaptive inference and compare with state-of-the-art-methods. We use average GFLOPs, a hardware independent metric, to measure the computation needed to classify all the videos in the testing set. We train \system with fixed $K$ time steps to obtain different models, denoted as \system-$K$ to accommodate different computational requirements during testing; and for each model we vary $\mu$ such that adaptive inference can be achieved within the same model.

\begin{figure}[t!]
\centering
\includegraphics[width=1.00\columnwidth]{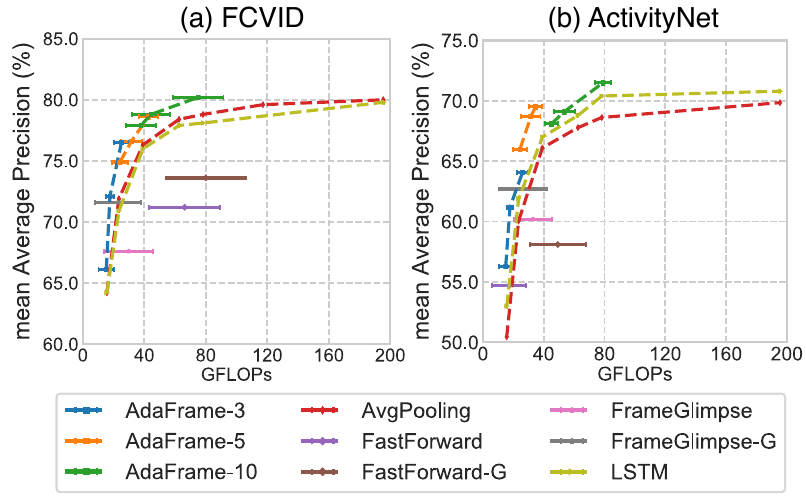}
\vspace{-0.2in}
\caption{ \textbf{Mean average precision \vs computational cost}. Comparisons of \system with FrameGlimpse~\cite{yeung2016end}, FastForward~\cite{fan18ijcai}, and alternative frame selection methods based on heuristics.} 
\vspace{-0.2in}
\label{fig:flops}
\end{figure}

In addition to selecting frames based on heuristics, we also compare \system with FrameGlimpse~\cite{yeung2016end} and FastForward~\cite{fan18ijcai}. FrameGlimpse is developed for action detection with a location network to select frames and a stop network to decide whether to stop; ground-truth boundaries of actions are used as feedback to estimate the quality of selected frames. For classification, there is no such ground-truth and thus we preserve the architecture of FrameGlimpse but use our reward function. FastForward~\cite{fan18ijcai} samples from a predefined action set, determining how many steps to go forward. It also consists of a stop branch to decide whether to stop. In addition, we also attach the global memory to these frameworks for fair comparisons, denoted as FrameGlimpse-G and FastForward-G, respectively. Figure~\ref{fig:flops} presents the results. For \textsc{AvgPooling} and \textsc{LSTM}, accuracies gradually increase when more computation (frames) is used and then become saturated. Note that the computational cost for video classification grows linearly with the number of frames used, as the most expensive operation is extracting features with CNNs. For ResNet-101 it needs 7.82 GFLOPs to compute features and for \system, it takes an extra 1.32 GFLOPs due to the computation in global memory. Therefore, we expect more savings from \system when more frames are used.

Compared with \textsc{AvgPooling} and \textsc{LSTM} using 25 frames, \system-10 achieves better results while requiring $58.9\%$ and $63.3\%$ less computation on average on \fcvid (80.2 \vs $\sim$195 GFLOPs~\footnote{195.5 GFLOPS for \textsc{AvgPooling} and 195.8 GFLOPs for \textsc{LSTM}.}) and \anet (71.5 \vs $\sim$195 GFLOPs), respectively. Similar trends can also be found for \system-5 and \system-3 on both datasets. While the computational saving of \system over \textsc{AvgPooling} and \textsc{LSTM} reduces when fewer frames are used, accuracies of \system are still clearly better, \ie, $66.1\%$ \vs $64.2\%$ on \fcvid, and $56.3\%$ \vs $53.0\%$ on \anet. Further, \system also outperforms FrameGlimpse~\cite{yeung2016end} and FastForward~\cite{fan18ijcai} that aim to learn frame usage by clear margins, demonstrating that coupling the training of frame selection and learning to stop with reinforcement learning on large-scale datasets without sufficient background videos is difficult. In addition, the use of a global memory, providing context information improves accuracies of the original model in both frameworks.
\begin{figure}[t!]
\centering
\includegraphics[width=1\columnwidth]{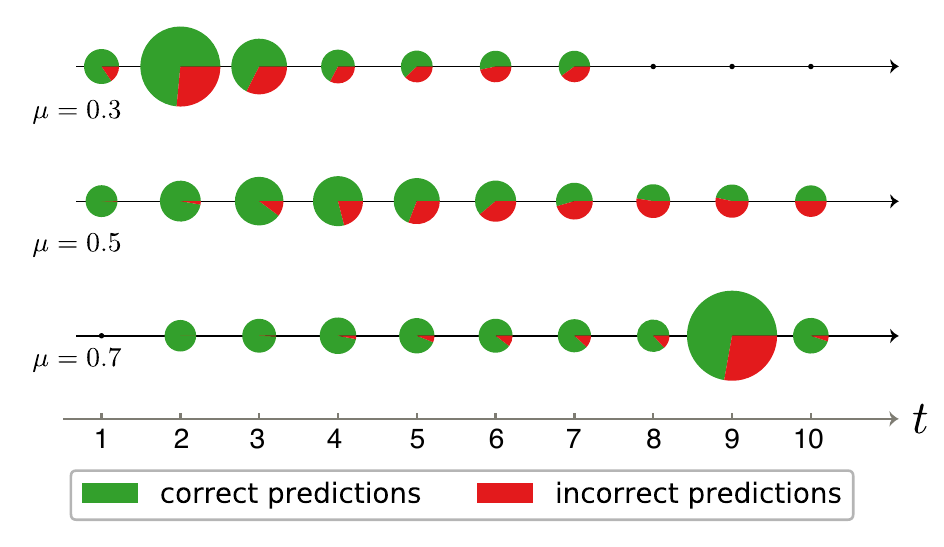}
\vspace{-0.15in}
\caption{ \textbf{Dataflow through \system over time}. Each circle represents, by size, the percentage of samples that are classified at the corresponding time step.} 
\label{fig:timestep}
\end{figure}

We can also see that changing the threshold $\mu$ within the same model can also adjust computation needed; the performance and average frame usage declines simultaneously as the threshold becomes smaller, forcing the model to make predictions as early as possible. But the resulting policies with different thresholds still outperform alternative counterparts in both accuracy and computation required. 

Comparing across different models of \system, we observe that the best model of \system trained with a smaller $K$ achieves better or comparable results over \system optimized with a large $K$ using a smaller threshold. For example, \system-3 with $\mu=0.7$ achieves an mAP of $76.5\%$ using 25.1 GFLOPs on \fcvid, which is better than \system-5 with $\mu=0.5$ that produces an mAP of $76.6\%$ with 31.6 GFLOPs on average. This possibly results from the discrepancies between training and testing---during training a large $K$ allows the model to ``ponder'' before emitting predictions. While computation can be adjusted with varying thresholds at test time, \system-10 is not fully optimized for classification with extremely limited information as is \system-3. This highlights the need to use different models based on computational requirements.

\begin{figure}[t!]
\centering
\includegraphics[width=1\columnwidth]{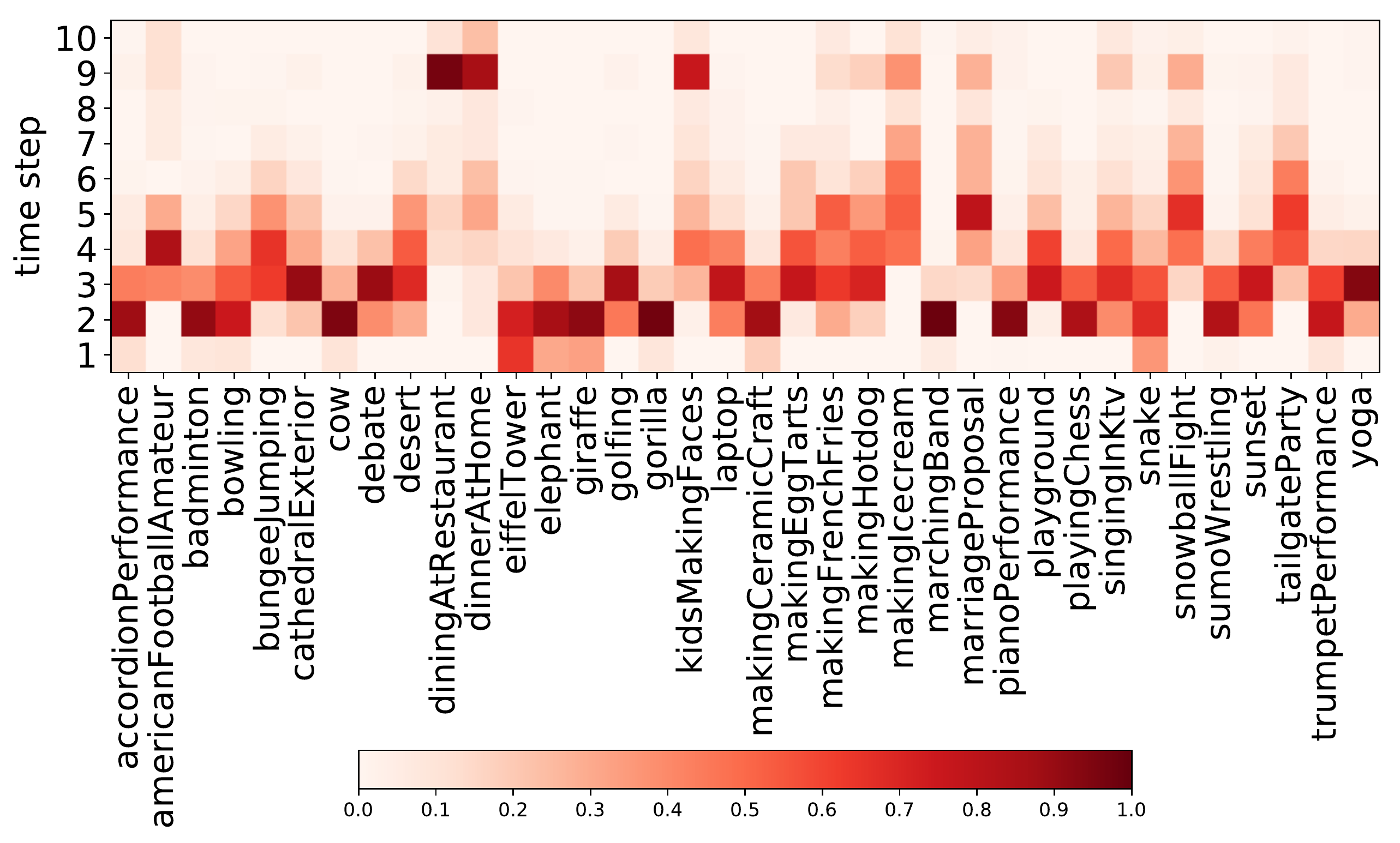}
\vspace{-0.1in}
\caption{\textbf{Learned inference policies for different classes over time.} Each square, by density, indicates the fraction of samples that are classified at the corresponding time step from a certain class in \fcvid.}
\label{fig:heatmap}
\vspace{-0.1in}
\end{figure}

\begin{figure*}[t!]
\centering
\includegraphics[width=0.9\linewidth]{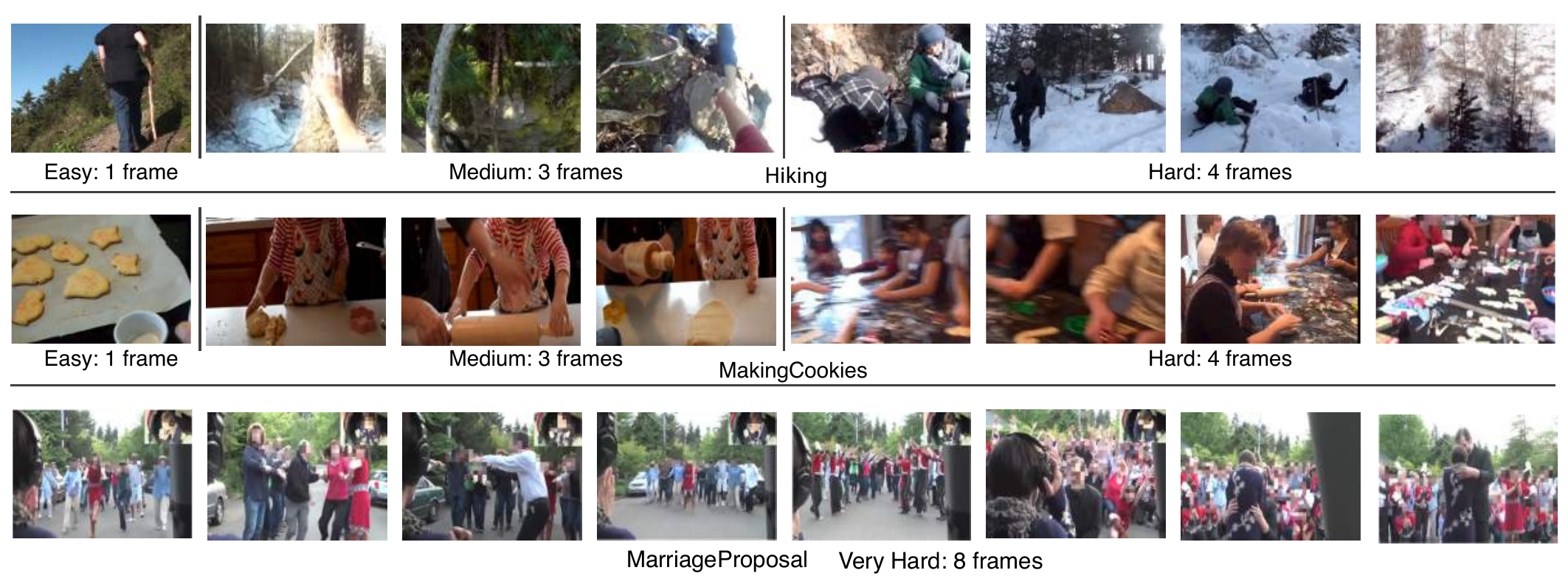}
\vspace{-0.12in}
\caption{\textbf{Validation videos from \fcvid using different number of frames for inference.}  Frame usage differs not only among different categories but also within the same class (\eg, ``making cookies'' and ``hiking'').}
\vspace{-0.16in}
\label{fig:difficulty}
\end{figure*}

\vspace{0.05in}
\noindent\textbf{Analyses of learned policies}. To gain a better understanding of what is learned in \system, we take the trained \system-10 model and vary the threshold to accommodate different computational needs. And we visualize in Figure~\ref{fig:timestep}, at each time step, how many samples are classified, and the prediction accuracies of these samples. We can see high prediction accuracies tend to appear in early time steps, pushing difficult decisions that require more scrutiny downstream. And more samples emit predictions at later time steps when computational budget increases (larger $\mu$). 

We further investigate whether computations vary for different categories. To this end, we show the fraction of samples from a subset of classes  in \fcvid that are classified at each time step in Figure~\ref{fig:heatmap}. We observe that, for simple classes like objects (\eg, ``gorilla'' and ``elephants'') and scenes (``Eiffel tower'' and ``cathedral exterior''), \system makes predictions for most of the samples in the first three steps; while for some complicated DIY categories (\eg, ``making ice cream'' and ``making egg tarts''), it tends to classify in the middle of the entire time horizon. In addition, \system takes additional time steps to differentiate very confusing classes like ``dining at restaurant'' and ``dining at home''. Figure~\ref{fig:difficulty} further illustrates samples using different numbers of frames for inference. 
We can see that frame usage varies not only across different classes but also within the same category (see the top two rows of Figure~\ref{fig:difficulty}) due to large intra-class variations. For example, for the ``making cookies'' category, it takes \system four steps to make correct predictions when the video contains severe camera motions and cluttered backgrounds. 

In addition, we also examine where the model jumps at each step; for \system-10 with $\mu=0.7$, we found that it goes backward at least once for $42.8\%$ of videos on \fcvid to re-examine past information instead of always going forward, confirming the flexibility \system enjoys when searching over time.
\subsection{Discussions}
In this section, we conduct a set of experiments to justify our design choices of \system.
\begin{table}[t!]
\centering
\small
\ra{0.8}
\resizebox{0.8\linewidth}{!}{
\begin{tabular}{@{}cc*{5}c@{}}
\toprule
\multicolumn{4}{c}{Global Memory} & \multicolumn{2}{c}{Inference} \\ 
\cmidrule{1-3} \cmidrule{5-6}
 & \# Frames & Overhead &	& mAP & \# Frames \\  
\cmidrule{1-3} \cmidrule{5-6}
& 0  & 0 && 77.9 & 8.40 \\ 
& 12 & 0.98  && 79.2 & 8.53\\
& 32 & 2.61  && 80.2 & 8.24\\ \midrule
& 16 & 1.32  & & \textbf{80.2} & \textbf{8.21} \\ 
\bottomrule
\end{tabular}}
\vspace{-0.1in}
\caption{\textbf{Results of using different global memories on \fcvid}. Different number of frames are used to generate different global memories. The overhead is measured for each frame compared to a standard ResNet-101.}
\label{tbl:global}
\end{table}

\vspace{0.05in}
\noindent\textbf{Global memory}. We perform an ablation study to see how many frames are needed in the global memory. Table~\ref{tbl:global} presents the results. The use of a global memory module improves the non-memory model with clear margins. In addition, we observe using 16 frames offers the best trade-off between computational overheads and accuracies.

\vspace{0.05in}
\noindent\textbf{Reward function}. Our reward function forces the model to increase its confidence when seeing more frames, to measure the transition from the last time step. We further compare with two reward functions:
\begin{enumerate*}[label=(\arabic*)]
\item \textsc{Prediction Reward}, that uses the prediction confidence of the ground-truth class $p_t^{gt}$ as reward;
\item \textsc{Prediction Transition Reward}, that uses $p^{gt}_{t} - p^{gt}_{t-1}$ as reward.
\end{enumerate*}
The results are summarized in Table~\ref{tbl:rewards}. We can see that our reward function and \textsc{Prediction Transition Reward}, both modeling prediction differences over time, outperform \textsc{Prediction Reward} that is simply based on predictions from the current step. This verifies that forcing the model to increase its confidence when viewing more frames can provide feedback about the quality of selected frames. Our result is also better than \textsc{Prediction Transition Reward} by further introducing a margin between predictions from the ground-truth class and other classes. 

\begin{table}[t!]
\centering
\small
\ra{0.9}
\resizebox{0.9\linewidth}{!}{
\begin{tabular}{@{}cc*{2}c@{}}
\toprule
	& Reward function	& mAP & \# Frames \\
	\cmidrule{2-4}
    & \textsc{Prediction Reward} &   78.7  & 8.34	 \\
    & \textsc{Prediction Transition Reward} & 78.9 & 8.31 \\ \midrule
           & Ours & \textbf{80.2} &  \textbf{8.21} \\
\bottomrule
\end{tabular}}
\vspace{-0.12in}
\caption{\textbf{Comparisons of different reward functions on \fcvid}. Frames used on average and the resulting mAP. }
\label{tbl:rewards}
\end{table}

\vspace{0.05in}
\noindent\textbf{Stop criterion}. In our framework, we use the predicted utility, measuring future rewards of seeing more frames, to decide whether to continue inference or not. An alternative is to simply rely on the entropy of predictions, as a proxy to measure the confidence of classifiers. We also experimented with entropy to stop inference, however we found that it cannot enable adaptive inference based on different thresholds. We observed that predictions over time are not as smooth as predicted utilities, \ie, high entropies in early steps and extremely low entropies in the last few steps. In contrast, utilities are computed to measure future rewards, explicitly considering future information from the very first step, which leads to smooth transitions over time. 

\section{Conclusion}
In this paper, we presented \system, an approach that derives an effective frame usage policy so as to use a small number of frames on a per-video basis with an aim to reduce the overall computational cost. It contains an LSTM network augmented with a global memory to inject global context information. \system is trained with policy gradient methods to predict which frame to use and calculate future utilities. During testing, we leverage the predicted utility for adaptive inference. Extensive results provide strong qualitative and quantitative evidence that \system can derive strong frame usage policies based on inputs.

\noindent{\footnotesize
\textbf{Acknowledgment}
ZW and LSD are supported by the Intelligence Advanced Research Projects Activity (IARPA) via
Department of Interior/Interior Business Center (DOI/IBC) contract number D17PC00345. The
U.S. Government is authorized to reproduce and distribute reprints for Governmental purposes not
withstanding any copyright annotation thereon. Disclaimer: The views and conclusions contained
herein are those of the authors and should not be interpreted as necessarily representing the official
policies or endorsements, either expressed or implied of IARPA, DOI/IBC or the U.S. Government.}

\bibliographystyle{ieee}
\bibliography{reference}

\end{document}